\documentclass[pre,aps,twocolumn,showpacs,showkeys]{revtex4-1}
\usepackage{graphicx}
\begin{document}
\title{Benford's Law and First Letter of Words}

\author{Xiaoyong Yan$^{1,2}$, Seong-Gyu Yang$^{3}$,  Beom Jun Kim$^{3}$ and Petter Minnhagen$^{4}$}
\email[]{Petter.Minnhagen@physics.umu.se}

\affiliation{$^1$Systems Science Institute, Beijing Jiaotong University, Beijing 100044, China\\
$^2$Big Data Research Center, University of Electronic Science and Technology of China, Chengdu 611731, China\\
$^3$Department of Physics, Sungkyunkwan University, Suwon 16419, Korea\\
$^4$IceLab, Department of Physics, Ume{\aa} University, 901 87 Ume{\aa}, Sweden
}

\begin{abstract}
A universal First-Letter Law (FLL) is derived and described. It predicts the percentages of first letters for words in novels. The FLL is akin to Benford's law (BL) of first digits, which predicts the percentages of first digits in a data collection of numbers. Both are universal in the sense that FLL only depends on the numbers of letters in the alphabet, whereas BL only depends on the number of digits in the base of the number system. The existence of these types of universal laws appears counter-intuitive. Nonetheless both describe data very well.
Relations to some earlier works are given. FLL predicts that an English author on the average starts about 16 out of 100 words with the English letter `\textit{t}'. This is corroborated by data, yet an author can freely write anything. Fuller implications and the applicability of FLL remain for the future.
\end{abstract}


\maketitle

\section{Introduction - Benford revisited}
\label{sec:sec1}

Benford's law predicts relative occurrence of first digits for numbers in a data collection~\cite{miller2015}.
The fact that a data collection of numbers often follows a specific universal distribution of first digits is rather counter-intuitive: it is sometimes unknown to persons involved in creating faked accounts and in such cases the law can be used to uncover such frauds~\cite{miller2015}.
The law can be expressed as 
\begin{equation}
p_i=\log_{X}\Bigl(\frac{i+1}{i}\Bigr),
\end{equation}
where $X$ is the base for the number system, $i=1,2,\cdots,X-1$ is the first digit of a number and $p_i$ the fraction of numbers which starts with the digit $i$. For example, the most common first digit is according to this law is 1 and has the relative frequency $p_1=\log_X 2$. For the usual decimal system with $X=10$ this gives $p_1=\log_{10}2\approx 0.301$, which means that 30.1\% of the numbers in the collection is predicted to start with the digit 1. If the numbers are instead given within the binary system with $X=2$, then $p_1=\log_2 2=1$, since in the binary system all numbers must start with the digit 1. 
 
\begin{figure}
\includegraphics[width=0.5\textwidth]{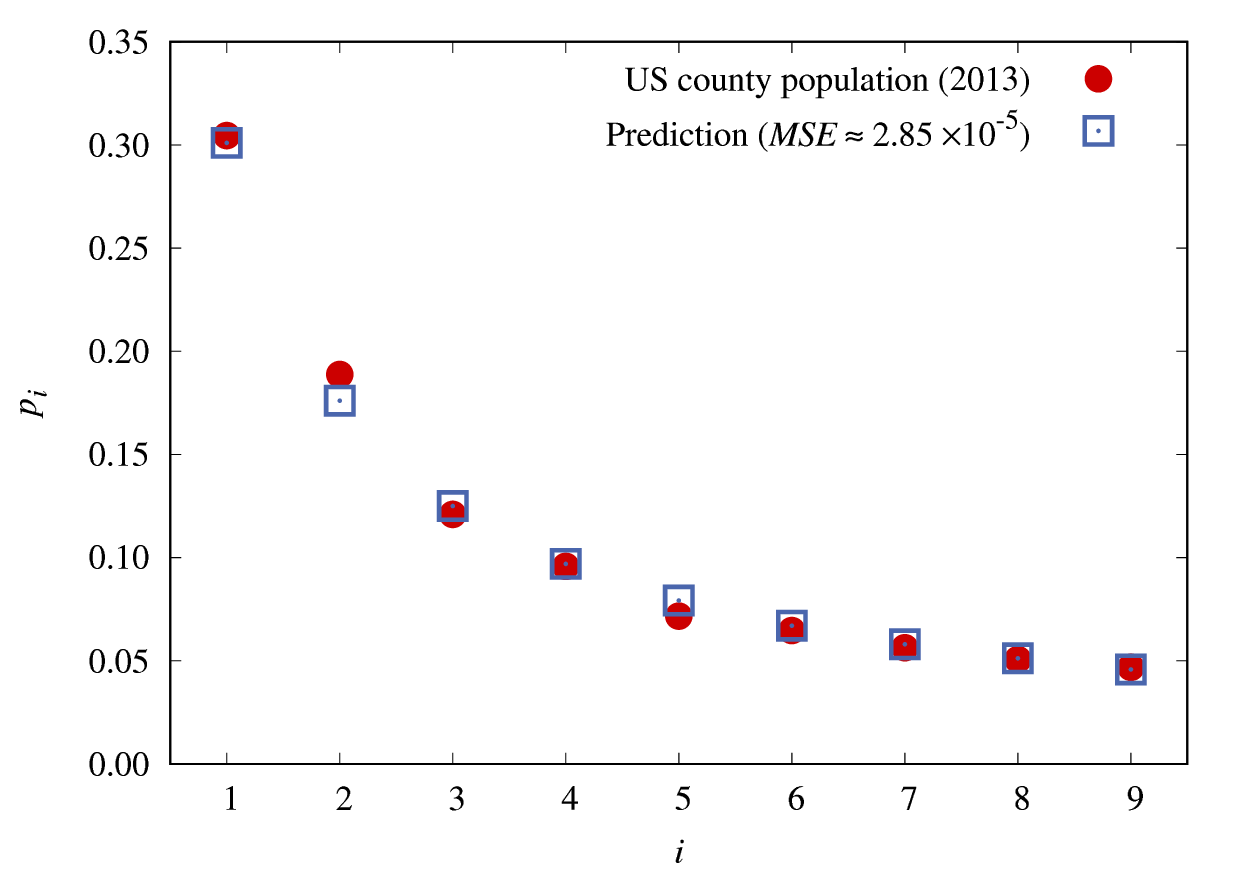}
\caption{
Distribution of first digits for the number of people belonging to US counties (filled circles, from US population census 2013~\cite{census}) compared to the Benford's law prediction (open squares). MSE (Mean Square Error between the data and the prediction) is $2.85\times10^{-5}$.
}
\label{fig1}
\end{figure}

The surprising thing is that the Benford's law to a good approximation is borne out by wide range of data collections from very different contexts~\cite{miller2015}. Figure~\ref{fig1} illustrates the Benford's law in the case of the size distribution of counties in USA. Mean Square Error (MSE) between the data and the Benford's law prediction is small which is about $2.85\times10^{-5}$ for 2013 US county-size distribution. The fact that Benford's law is borne out by a wide range of systems, and that the law carries absolutely no specific information on the system itself, compared with the fact that it has no free parameters, points to a general origin~\cite{miller2015}.

Benford independently made his discovery around 1938~\cite{benford1938}. However, the law had been discovered earlier by Newcomb around 1880~\cite{newcomb1881}. Newcomb's discovery allegedly originated from his observation that the logarithm table in a public library used by many, was more worn in the beginning, where people had been looking for logarithm for numbers starting with 1 and less at the end, which gave the logarithms for numbers starting with 9. Thus, the numbers looked up by the users of this logarithmic table were in fact collected into groups labeled by the starting digit. The relation between the various group sizes should, according to Newcomb, be given by Benford's law~\cite{newcomb1881}.

Newcomb's observation ties into the present investigation in the following way: Suppose that, instead of looking up logarithms in a table of logarithms, you were looking up words in a dictionary. Suppose you were reading the novel {\it Moby Dick} by Herman Melville and that you did not know English, so that you had to look up every word in a dictionary into your native language. Furthermore assume that your memory was so bad that you had to look up every word each time occurred in the text. This would mean that you effectively collected words into the first letter groups. These groups would be in different sizes just as the first digit groups for numbers: you would look up more times in the dictionary for certain starting letters than for others. One may then ask if there is some law, akin to Benford's law, which gives the distribution of the group sizes for the first letters. 

The First-Letter Law (FLL), derived and discussed in the present paper, is such a law. It can be expressed as

\begin{equation}
p_i=\frac{X-(X-1)\log_{X} (X-1) - i\log_{X} i+(i-1)\log_{X}(i-1)}{ X(X-1)\log_X\bigl({\frac{X}{X-1}}\bigr)}.
\label{alaw}
\end{equation}
Here $X (\geq 2)$ is the number of letters in the alphabet, $p_i$ is the ratio of the first letter group $i$ where $i=1$ is the most frequent first letter and $i=X$ is the least frequent one. According to our First-Letter Law, Eq.~(\ref{alaw}), the ratio of the most frequent first letter for an English novel is
\begin{equation}
p_1=\frac{1+25\log_{26} \bigl({\frac{26}{25} \bigr)}}{26\cdot 25\log_{26}\bigl({\frac{26}{25}}\bigr)},
\end{equation}
since the English alphabet has $X=26$ letters. This gives $p_1\approx 0.166$. Table~\ref{table:percentage} gives FLL-prediction for a 26-letter alphabet as the percentage for the occurrence of a first letter of a word in terms of its rank ($i=1$: the most common first letter, $i=2$: the second most common, and so on).

\begin{table}
\label{table:percentage}
\centering
\scalebox{0.9}{
\begin{tabular}{ |c|c|c|c|c|c|c|c|c|c|c|c|c| }
\cline{1-6}\cline{8-13}
$i$&\bf{FLL}&\bf{MD}&\bf{TU}&\bf{MS}&\bf{AN}&\quad &$i$&\bf{FLL}&\bf{MD}&\bf{TU}&\bf{MS}&\bf{AN}\\
\cline{1-6}\cline{8-13}
1 &\bf{16.6}&16.4&16.3&15.9&\it{16.1}& &14	&\bf{2.5}&2.2&2.5&2.9&\it{2.6}\\
2 &\bf{11.2}&10.9&10.7&11.7&\it{11.7}& &15	&\bf{2.2}&1.8&2.5&2.4&\it{2.4}\\
3 &\bf{9.1} &9.3 &8.5 &9.2 &\it{8.6} & &16	&\bf{2.0}&1.8&2.0&2.3&\it{2.2}\\
4 &\bf{7.8} &6.7 &8.3 &7.3 &\it{7.3} & &17	&\bf{1.7}&1.8&1.8&2.0&\it{2.2}\\
5 &\bf{6.8} &6.7 &7.3 &6.8 &\it{7.2} & &18	&\bf{1.5}&1.6&1.7&1.9&\it{1.7}\\
6 &\bf{6.0} &6.5 &6.1 &5.8 &\it{5.9} & &19	&\bf{1.3}&1.2&1.5&1.7&\it{1.7}\\
7 &\bf{5.4} &6.4 &6.0 &4.7 &\it{5.1} & &20	&\bf{1.1}&1.2&1.2&1.0&\it{1.2}\\
8 &\bf{4.8} &5.2 &4.4 &4.5 &\it{4.6} & &21	&\bf{0.9}&0.7&0.6&0.7&\it{0.7}\\
9 &\bf{4.3} &3.9 &3.9 &4.1 &\it{3.9} & &22	&\bf{0.7}&0.4&0.6&0.6&\it{0.5}\\
10&\bf{3.9} &3.9 &3.7 &3.7 &\it{3.8} & &23	&\bf{0.5}&0.4&0.3&0.4&\it{0.4}\\
11&\bf{3.5} &2.8 &3.7 &3.7 &\it{3.6} & &24	&\bf{0.3}&0.3&0.2&0.2&\it{0.2}\\
12&\bf{3.1} &2.8 &3.4 &3.5 &\it{3.5} & &25	&\bf{0.2}&0.0&0.0&0.0&\it{0.0}\\
13&\bf{2.8} &2.7 &2.7 &2.9 &\it{3.0} & &26	&\bf{0.0}&0.0&0.0&0.0&\it{0.0}\\
\cline{1-6}\cline{8-13}
\end{tabular}}

\caption{
The table gives first letter percentages for five cases in five column: Column \textbf{FLL} gives the predictions from FLL in bold-face; \textbf{MD} the percentages from {\it Moby Dick}, \textbf{TU} the percentages \textit{Tess of the d'Urbervilles} (see Fig.~\ref{fig3}); \textbf{MS} the percentages for \textit{Main Street} (see Fig.~\ref{fig3}); \textbf{AN} the average percentage for the nine novels in Fig.~\ref{fig3} (see Fig.~\ref{fig4}).
}
\end{table}

\begin{figure}
\includegraphics[width=0.5\textwidth]{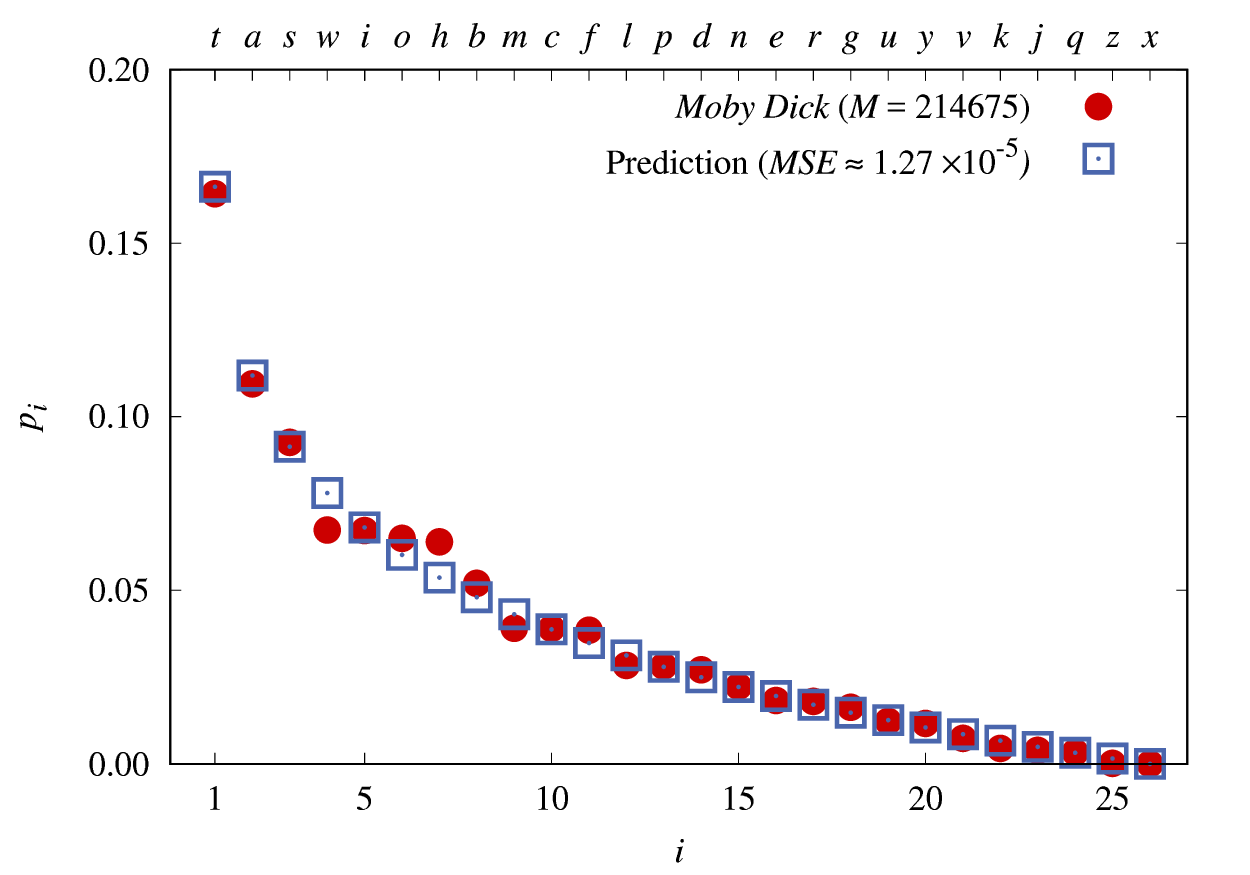}
\caption{
Distribution of the first letter of words for the novel {\it Moby Dick} by Herman Melville (filled circles) compared to the FLL given by Eq.~(\ref{alaw}) (open squares).
The agreement is striking (compare with Table~\ref{table:percentage}, the MSE is $1.27 \times 10^{-5}$). {\it Moby Dick} contains a total of $M=214675$ words. The actual first letters are given at the top (`{\it t}', `{\it a}', `{\it s}', `{\it w}',$\cdots$, `{\it q}', `{\it z}', `{\it x}').
}
\label{fig2}
\end{figure}

Figure~\ref{fig2} shows the validity of our FLL for the case of the novel {\it Moby Dick}. The overall agreement is strikingly good and the most frequent first letter has $p_1= 0.164$ (which is the letter `\textit{t}'), very close to the predicted ratio 0.166. Table~\ref{table:percentage} gives all percentages for the first letters in {\it Moby Dick} and other novels.

The structural similarity between Benford's law (BL) and FLL is that neither of them contains any free parameter: BL only has the number of digits in the numerical base as its input and FLL only the number of letters in the alphabet.

Neither BL nor FLL are always valid: Benford's law has been thoroughly investigated with multitude of examples. Its limitations have been established and possible ways of understanding its origin have been proposed and debated~\cite{miller2015,pietronero2001,whyman2016,fewster2009}.
The scope of the present investigation is more narrow and more in the spirit of Newcomb's original discovery of sifting first digits of numbers into groups~\cite{newcomb1881}. A difference between Newcomb's discovery and ours is that Newcomb first found the evidence in the data and from this deduced BL, whereas we first arrived at the FLL given by Eq.~(\ref{alaw}) from the arguments given below and then compared to data. 

In the following Section~\ref{sec:sec2} we present and discuss evidences for the FLL. Section~\ref{sec:sec3} gives our explanation for the existence of the law with some details relegated to Appendix~\ref{app:appA}. The reason that leads to FLL is explained in Section~\ref{sec:sec4} with details described in Appendix~\ref{app:appB}. Finally, Section~\ref{sec:sec5} gives some more perspectives and some outlook.

\section{Evidence of First-Letter Law}
\label{sec:sec2}

\begin{figure*}
\includegraphics[width=1\textwidth]{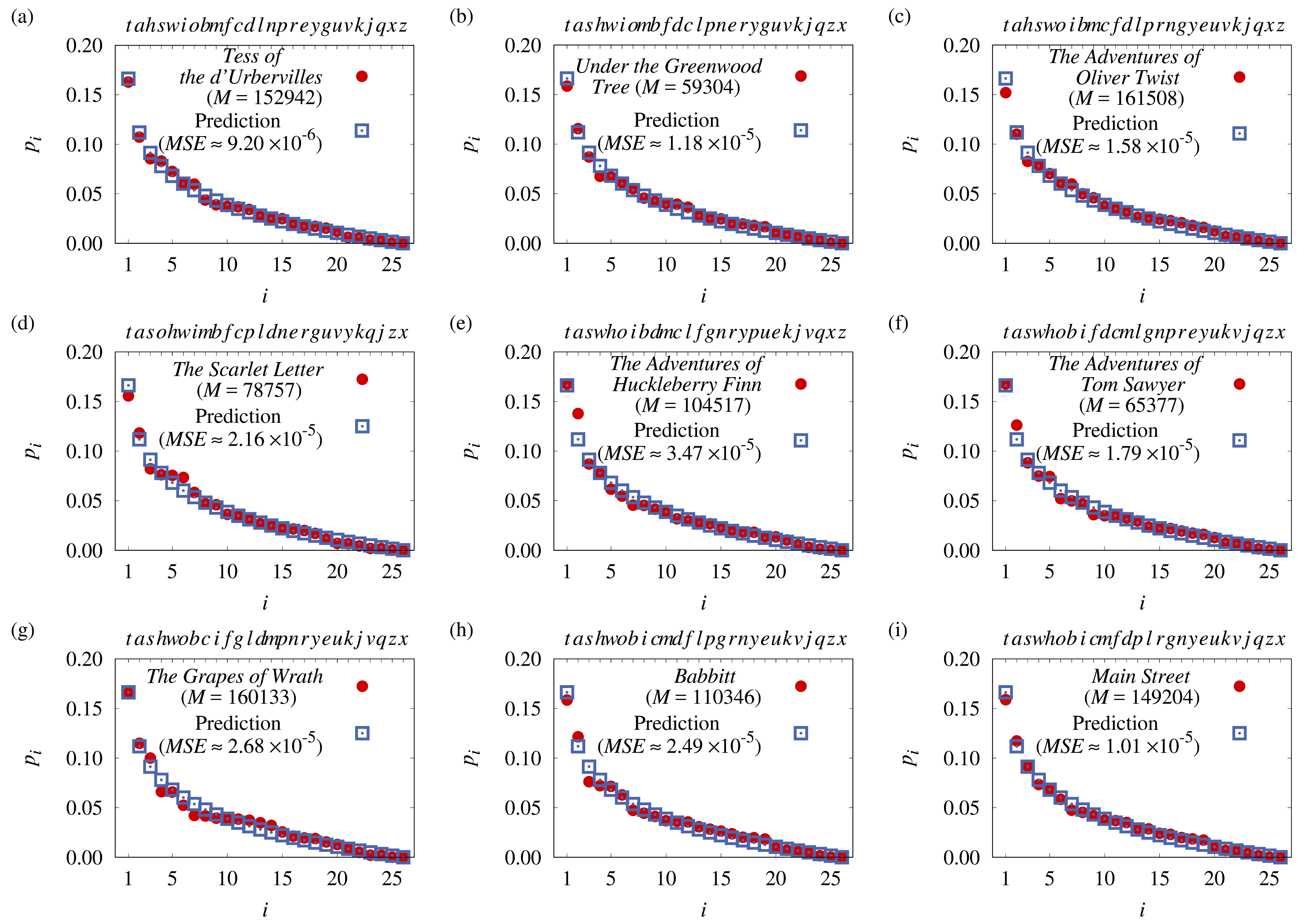}
\caption{
Comparison between FLL (open squares) and nine novels written in English by six authors (filled circles). For each novel the total number of words $M$ and the MSE values are given in the panels and the actual first letters at the top of the panels. The MSE values range from about $0.9\times 10^{-5}$ to $3.5\times 10^{-5}$. The novels are: (a) \textit{Tess of d'Urbervilles} (see Table~\ref{table:percentage} for actual numbers), (b) \textit{Under the Greenwood tree} by Thomas Hardy; (c) The \textit{Adventures of Oliver Twist} by Charles Dickens; (d) \textit{The Scarlet Letter} by Nathaniel Hawthorne; (e) \textit{The Adventures of Huckleberry Finn}, (f) \textit{The Adventures of Tom Sawyer} by Mark Twain; (g) \textit{The Grapes of Wrath} by John Steinbeck; (h) \textit{Babbitt} and (i) \textit{Main Street} by Sinclair Lewis (see also Table~\ref{table:percentage} for actual numbers).
} 
\label{fig3}
\end{figure*}

Example of the FLL can be found in classical novels written by well-known English and American authors. Figure~\ref{fig3} illustrates this by nine such novels~\cite{gutenberg}. The first row is for the novels \textit{Tess of the d'Urbervilles} (1891) and \textit{Under the Greenwood Tree} (1872) by Thomas Hardy and \textit{The Adventures of Oliver Twist} (1837) by Charles Dickens. Hardy and Dickens were English authors. The next row represents three American novels from roughly the same time: \textit{The Scarlet Letter} (1850) by Nathaniel Hawthorne, \textit{The Adventures of Huckleberry Finn} (1884) and \textit{Tom Sawyer} (1884) by Mark Twain. Finally, the lowest row represents three somewhat more recent American novels: \textit{The Grapes of Wrath} (1939) by John Steinbeck, \textit{Babbitt} (1920) and \textit{Main Street} (1920) by Sinclair Lewis. In each case the total number $M$ of words is given as well as the MSE between the data and the FLL-prediction. At the top the actual first letters are given for each novel. Table~\ref{table:percentage} lists all the percentages for the first and the last novels (\textit{Tess of the d'Urbervilles} and \textit{Main Street}). 

Figure~\ref{fig3} illustrates two crucial points: the first is that the actual frequency ratios for the first letters of words are very closely the same in all cases: in all cases the frequency ratios closely follow a unique frequency ladder. The surprise is that each author is creatively writing independently in his or her personal style developing very different themes. So what enforces the author in such a situation to prefer that close to 16\% out of all the words in the novel should start on `{\it t}' must originate from the same cause. The existence of a universal frequency ladder is even more surprising since the assignment of the actual first letters to a particular ladder-step differs somewhat from case to case. For example compare Fig.~\ref{fig3}(a) and (i), in which both overlap very well with the FLL and hence with each other. The third most frequent first letter is `\textit{h}' in Fig.~\ref{fig3}(a) but `\textit{s}' in (i) (compare with Table~\ref{table:percentage}). Even novels written by the same author may differ in the assignment of ladder steps: \textit{Tess of the d'Urbervilles} [Fig.~\ref{fig3}(a)] and \textit{Under the Greenwood Tree} [Fig.~\ref{fig3}(b)] were both written by Thomas Hardy, yet the assignment of the third largest ladder step differs. It is `{\it h}' in Fig.~\ref{fig3}(a) and `{\it s}' in Fig.~\ref{fig3}(b).

The second crucial point is that all the first letter frequencies in Fig.~\ref{fig3} are very well described by FLL. This is surprising since FLL only depends on the number of letters in the alphabet. Why would the number of letters in the alphabet constrain an English author to on the average start 16 out of 100 words with a `{\it t}'?

\begin{figure}
\includegraphics[width=0.5\textwidth]{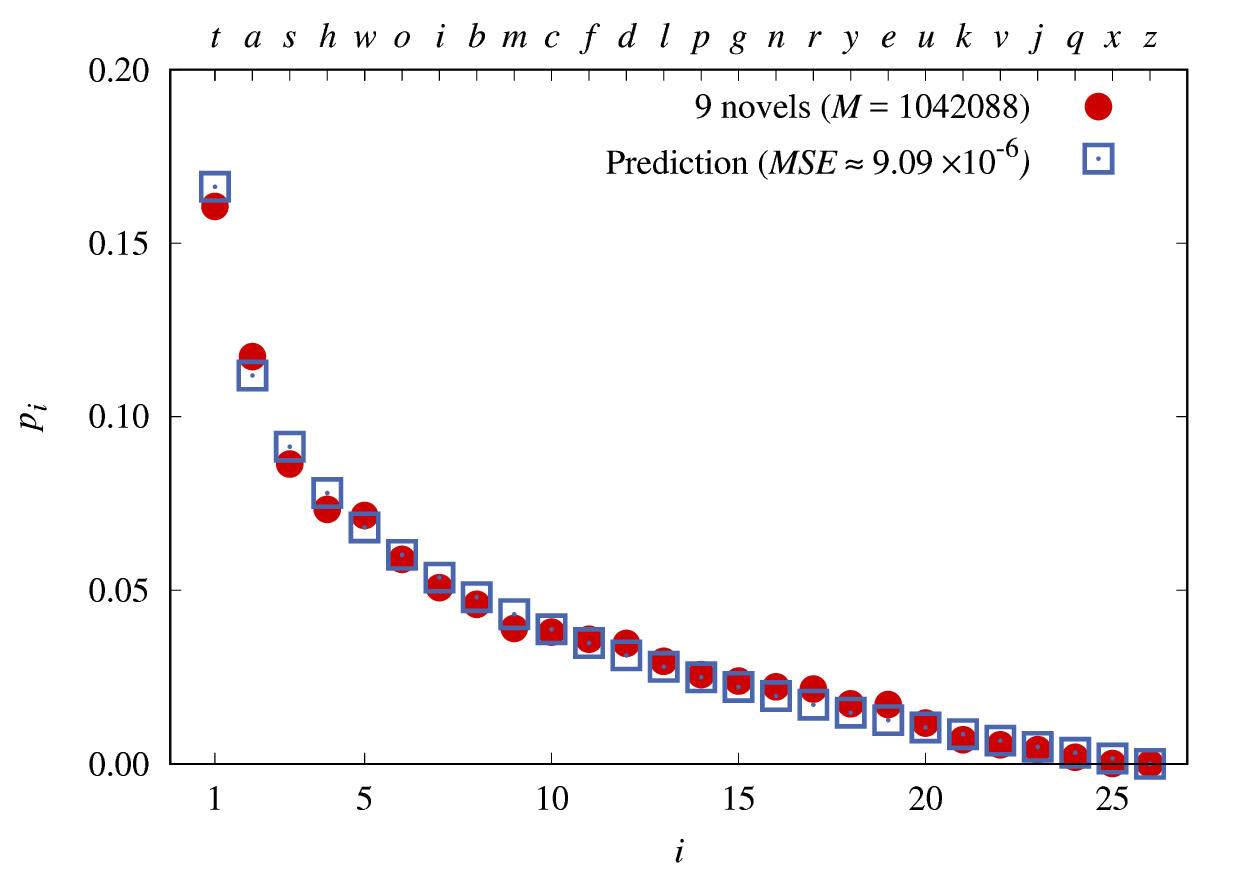}
\caption{
Average ratios for the nine novels shown in Fig.~\ref{fig3}. The averages are weighted by the relative sizes $M$ of the novels; Average ratios (filled circles) compared to FLL-predictions (open squares). The agreement is excellent (compare with Table~\ref{table:percentage}, the MSE is $9.09 \times 10^{-6}$). The nine novels contains in total $M=1042088$ words. The actual first letters are given at the top. The explicit numbers are given in Table~\ref{table:percentage}.
}
\label{fig4}
\end{figure}

Note that in Fig.~\ref{fig3}(a) \textit{Tess of d'Urbervilles} and Fig.~\ref{fig3}(i) \textit{Main Street} have the smallest MSE and also visually overlap excellently with the FLL (compare with Table~\ref{table:percentage}). \textit{The Adventures of Huckleberry Finn} has the largest MSE, yet the visual overlap with FLL is still quite impressive. Figure~\ref{fig4} shows the average of the ratio for all the nine novels in Fig.~\ref{fig3}. The explicit values are given in Table~\ref{table:percentage}: the agreement with FLL is even better for this average than for the individual novels. This suggests that the statistical error is decreased (see section~\ref{sec:sec3}).

\begin{figure}
\includegraphics[width=0.5\textwidth]{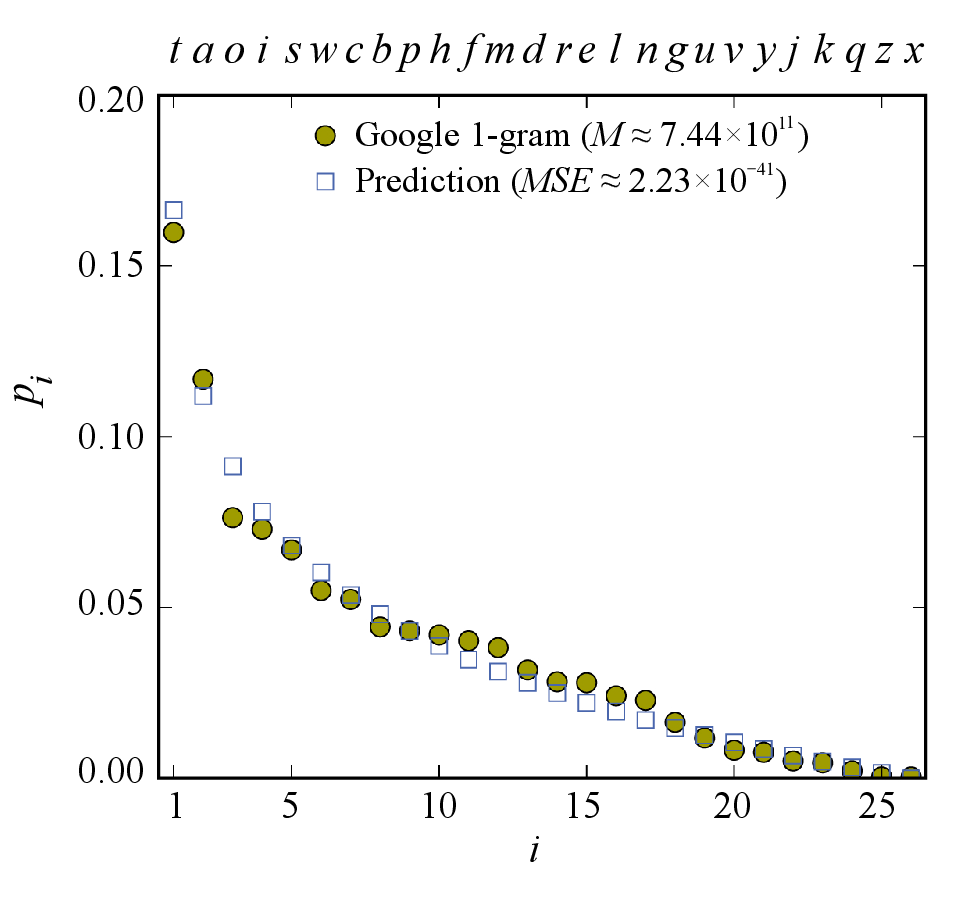}
\caption{
First letter distribution for a large collection of English texts (English Google 1-gram~\cite{google}). The number of words in this collection is $M\approx 7.4\times 10^{11}$ and hence roughly corresponds to millions of novels merged into one. The Google 1-gram data (filled circles) are compared to the FLL-prediction (open squares). The agreement is good (MSE $2.2 \times 10^{-5}$ which is close but slightly larger than the weighted average MSE $0.91 \times 10^{-5}$ for the nine novels in Fig.~\ref{fig3}).
}
\label{fig5}
\end{figure}

But it is possible that the data in Fig.~\ref{fig2} and Fig.~\ref{fig3} are not representative. One way of checking this is to use Google 1-gram which gives all the words contained in a large collection of English texts~\cite{google}. So whereas a typical novel contains a total of about $M=10^5$ words the Google 1-gram contains about $M=10^{12}$, roughly corresponding to $10^7$ English novels. Figure~\ref{fig5} shows that the universal frequency ladder is borne out to a good approximation also for this huge collection of English texts. This indicates that the data for the novels in Fig.~\ref{fig2} and Fig.~\ref{fig3} are indeed representative in the sense that many English novels do follow the universal frequency ladder of first letters.

\begin{figure}
\includegraphics[width=0.5\textwidth]{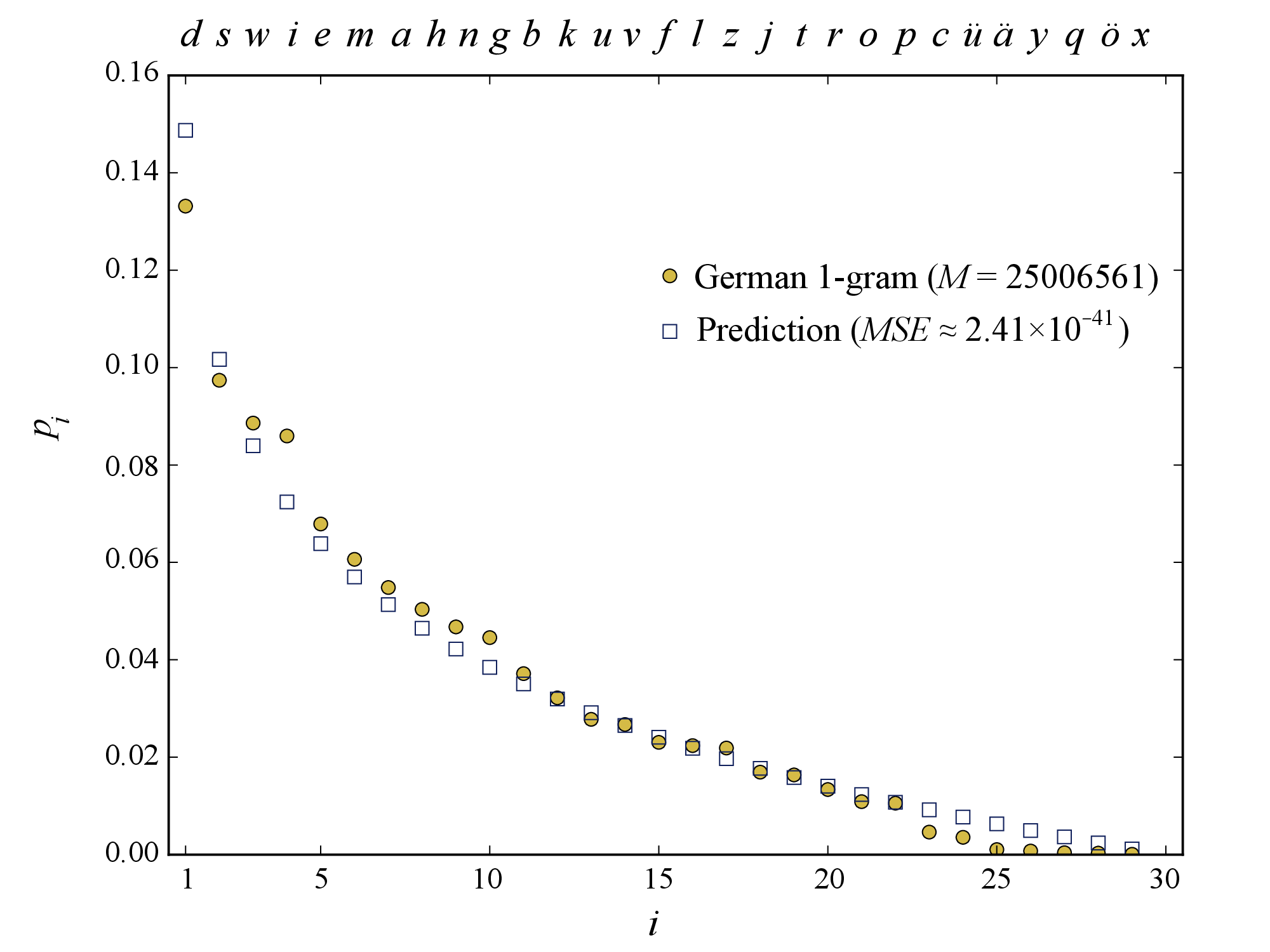}
\caption{
First letter distribution for a large collection of German texts (German Google 1-gram~\cite{google}). In this case the agreement with the German 1-gram data (filled circles) and FLL (open squares) is almost as good as for English 1-gram (MSE=$2.41 \times 10^{-5}$ for German and $2.23 \times 10^{-5}$ for English), but the agreement appears to be visually slightly better for English 1-gram.
}
\label{fig6}
\end{figure}

Exactly which novels follow and which do not follow FLL may depend on many factors. One of these factors is presumably the language. Figure~\ref{fig6} gives the frequency ladder for German 1-grams~\cite{google}. In this case the German alphabet has $X=30$ which means that the FLL-prediction Eq.~(\ref{alaw}) for the frequency ladder is slightly different. For example, the largest frequency is predicted to be 14.8\%. There is still a good agreement and MSE is just as good as for English 1-gram (The MSE=$2.41 \times 10^{-5}$ is to be compared to $2.23\times 10^{-5}$ for English 1-gram). This indicates that the FLL is no exclusive for English novels, but the quality of agreement may be better for some languages than for others. There might also be differences between various branches of English like differences between British and American English.

However, the intriguing question is the possible origin of the FLL, which we now turn our attention to.

\section{Origin of First-Letter Law}
\label{sec:sec3}

\begin{figure*}
\includegraphics[width=1\textwidth]{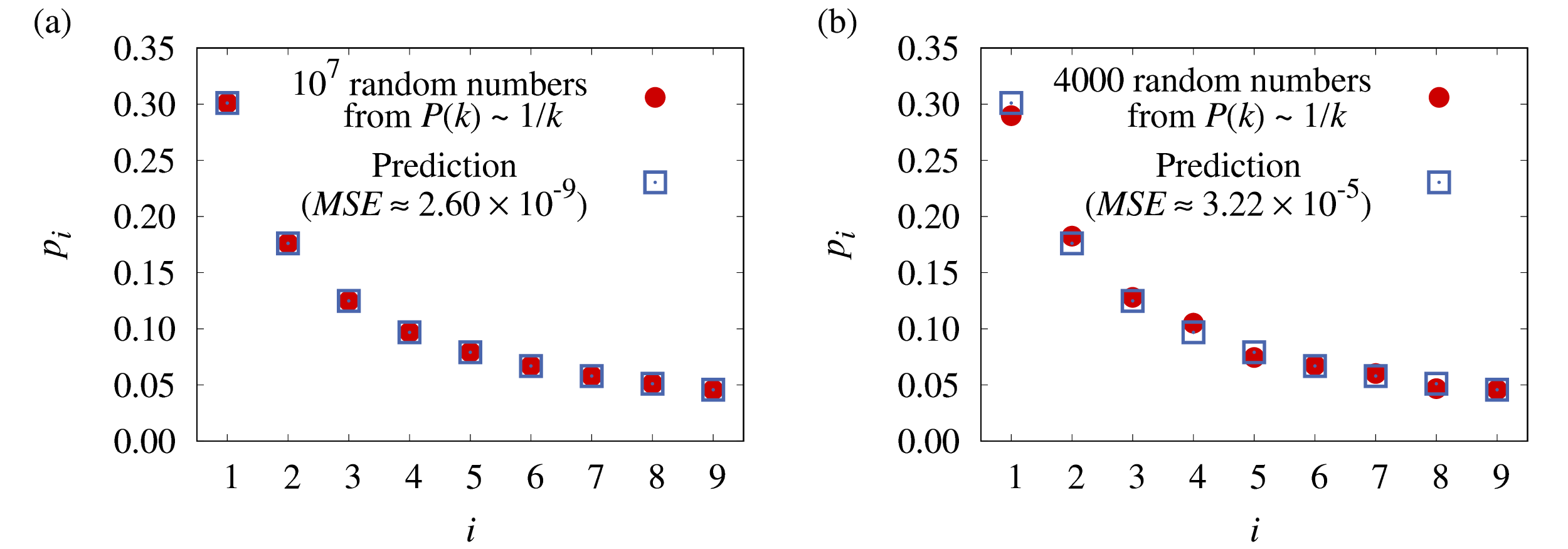}
\caption{
Illustration of how Benford's law is connected to \textit{randomly} drawing numbers from the probability distribution $P(k)\propto 1/k $. The more numbers you draw, the better becomes the agreement. (a) shows the result from randomly drawing $10^7$ numbers. In this case the overlap is almost perfect with and MSE about $10^{-9}$ and (b) shows that if you instead draw only 4000 numbers the statistical fluctuations increases the MSE to about $10^{-5}$.
}
\label{fig7}
\end{figure*}

In case of the Benford's law the origin is much investigated~\cite{miller2015}. One argument leading to the law is based on \emph{randomness}: You assume that your dataset of numbers is approximately equivalent to randomly drawn numbers $k$ from a probability distribution $P(k)$. The law is then a consequence of the property of this distribution. For the Benford's law to be \emph{precisely} (as opposed to approximately) valid, $P(k)$ must have a special form: All numbers $k$ in the interval $[10^{\log_{10}(i)}, 10^{\log_{10}(i+1)}]$ starts on the digit $i$ and the probability for picking a number $k$ in this interval is
\begin{equation}
p_i\propto \int_{\log_{10}i}^{\log_{10}(i+1)}kP(k) d\log_{10}k.
\end{equation}
 For $P(k)\propto 1/k$, the probability for picking the digit $i$ becomes $p_i=\log_{10}(\frac{i+1}{i})$ which is recognized as the crucial element in the Benford's law. This explanation of picking numbers randomly from $P(k)\propto 1/k$ is at the core of many explanations of the BL~\cite{fewster2009}.
Figure~\ref{fig7} shows that BL is indeed obtained by picking numbers randomly from this distribution. The more numbers you pick the smaller becomes the deviation between the data and the BL-prediction. Figure~\ref{fig7}(a) shows that a large number of picked numbers makes the BL-prediction precise whereas Fig.~\ref{fig7}(b) shows that picking fewer numbers results in a statistical random error between the BL-prediction and the data. This means that, according to this explanation of the BL, the deviation between BL and the data in general has two sources: the statistical error and the error caused by the fact that the numbers are not randomly picked from $P(k)\propto 1/k$.

In short, one explanation of BL is based on an assumed equivalence to randomly picking numbers from a probability distribution $P(k)$ of a form which approximately leads to Benford's law~\cite{miller2015,fewster2009}.

\begin{figure*}
\includegraphics[width=1\textwidth]{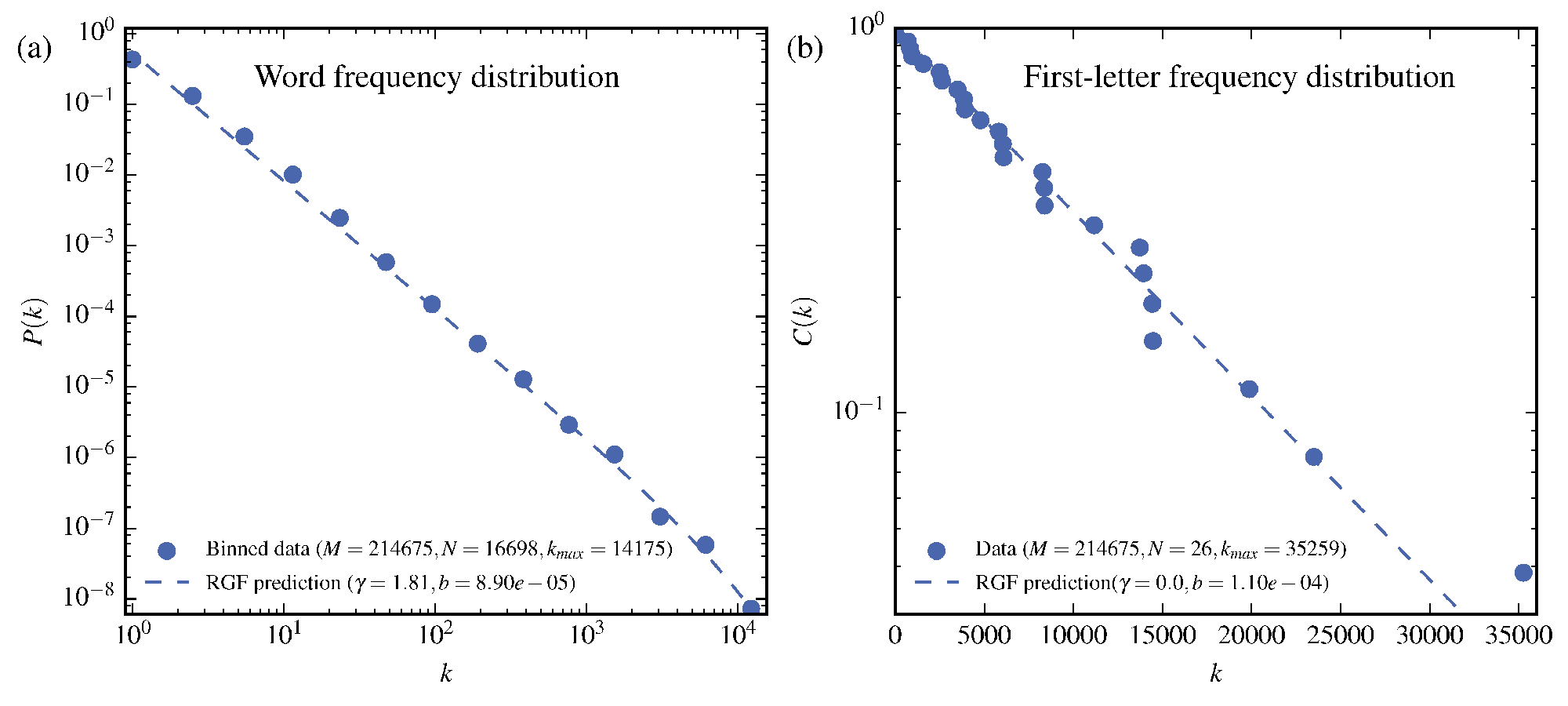}
\caption{
(a) Word frequency distribution for {\it Moby Dick} compared to the Random-Group Formation (RGF)-theory (see Appendix~\ref{app:appB} and Ref.~\cite{yan17}). (b) First-letter distribution compared to the RGF-prediction.
}
\label{fig8}
\end{figure*}

Our explanation of the First-Letter Law also follows from randomly drawing objects from a probability function of a particular form. To this end one needs to know the connection between a written text and randomness~\cite{baek11,yan15,yan15b,yan16,yan17}. In what sense are the words written by Herman Melville in the novel {\it Moby Dick} randomly drawn from a probability distribution? The novel {\it Moby Dick} contains in total $M$ words of which $N$ are different words [compare with Fig.~\ref{fig8}(a)]. Each different word appears a certain number of times in the text. The words can then be grouped into frequency groups such that $N(k)$ different words occur $k$ times in the text. The words contained in a frequency group are uniformly spread within the text. Thus from the perspective of frequency groups the process of writing the novel is equivalent to randomly picking words from a frequency group with the probability distribution $P(k)=N(k)/N$. This probability distribution is shown in Fig.~\ref{fig8}(a). In Ref.~\cite{baek11} it was shown that randomly grouping $M$ labeled objects into $N$ groups by a maximum entropy argument always leads to a $P(k)\propto \exp(-bk)/k^\gamma$ where $\gamma$ and $b$ are two constant uniquely determined from the average frequency $\langle k\rangle=M/N$ and the frequency $k_{\rm max}$ for the most frequent word (which is `{\it the}' in English). Figure~\ref{fig8}(a) shows that the frequency distribution for {\it Moby Dick} falls into this class. Thus the shape of word-frequency distribution is in fact determined by the randomness through the maximum entropy argument~\cite{baek11,yan15,yan15b,yan16,yan17}.

Suppose now that only the first letter of the words in {\it Moby Dick} are kept. This changes $N$ to $N=26$ but M remains the same. So $\langle k\rangle$ changes to $M/26$ and the most frequent letter `{\it t}' is now 16.4\% percent of the total number of words. The frequency distribution of the first letters of {\it Moby Dick} together with the expected distribution from random grouping are given in Fig.~\ref{fig8}(b). The random-group-prediction is close to an exponential, $P(k) \propto \exp(-bk)$ and agrees very well with the actual data. The First-Letter Law follows when the $P(k)$ is an exponential. The derivation is given in Appendix~\ref{app:appA}. This leaves us with the question as to why it is an exponential. 
\section{Why Exponential?}
\label{sec:sec4}

Based on our earlier work we here give a qualitative argument as to why the First-Letter Law is connected to an exponential probability distribution~\cite{baek11,yan15,yan15b,yan16,yan17}. A more quantitative argument, based on the results in Ref.~\cite{yan17}, is given in Appendix~\ref{app:appB}. Suppose you randomly group $M$ labeled objects into $N$ groups without any additional constraint. The information needed to identify a labeled object in a group of $k$ object is $\log_2 k$ and within the random group approach this information leads to the distribution resulting in $P(k)\propto \exp(-bk)/k$~\cite{baek11}. However, if you assemble the pieces for an IKEA furniture which should be put together at a particular joint containing $k$ pieces, then it is not enough to distinguish the parts. In addition you must distinguish in which order they should be joined together. So the information needed to succeed is this time $\log_2 k+\log_2 k=2\log_2 k$ (i.e., information for the correct piece + information for correct order). This leads to $P(k)\propto \exp(-bk)/k^2$~\cite{lee12}. Finally if neither order nor identity of the pieces matters the information required is $\log_2 1=0$ and the distribution becomes an exponential. 

The information $\gamma \log_2 k$ with $\gamma=$ 0, 1 and 2 represents three cases where the grouped objects carry different amount of effective distinguishable information. The limit case with $\gamma=0$ means that the effective distinguishable information is zero. This is equivalent to the grouping of indistinguishable objects and corresponds to an exponential $P(k)$. The words in {\it Moby Dick} corresponds to an effective distinguishable information $\gamma \log_2 k$ with $\gamma\approx 1.81$ [compare with Fig.~\ref{fig8}(a)]. However, as shown in Fig.~\ref{fig8}(b) the First-Letter distribution corresponds to good approximation to an exponential. This indicates that for the first letters $\gamma\approx 0$, suggesting a total loss of distinguishable information. You will clearly lose information when a word is replaced by its first letter. Or put it another way, you clearly must supply a lot of information in order to reconstruct the content of a novel if you only know the first letters.
 
Thus {\it Moby Dick} represents a case when you lose almost all information by deleting all except the first letters in the words. As a consequence the distribution of first letters for {\it Moby Dick} becomes exponential (see Appendix~\ref{app:appB} for more details). 

\section{Outlook}
\label{sec:sec5}

The FLL given by Eq.~(\ref{alaw}) was obtained as the maximum entropy solution within a Random-Group Formation(RGF) approach. It describes the situation when $M$ objects are randomly sorted within $N$ groups in such a way that the average number of objects per group is constant. From this the average sizes of the groups are calculated. FLL only gives the sizes of the groups, e.g., it predicts that the largest group contains 16.6\% of the objects, but it does not tell you that this particular group corresponds to the first letter `\textit{t}'. This means that the size ordering in terms of letters varies somewhat, e.g., the third largest group in our examples are most often `{\it s}' but sometimes `{\it h}' and `{\it o}'. This is different from the Benford's law which \textit{a priori} identifies the size ordering directly with the first digit.

\begin{figure*}
\includegraphics[width=1\textwidth]{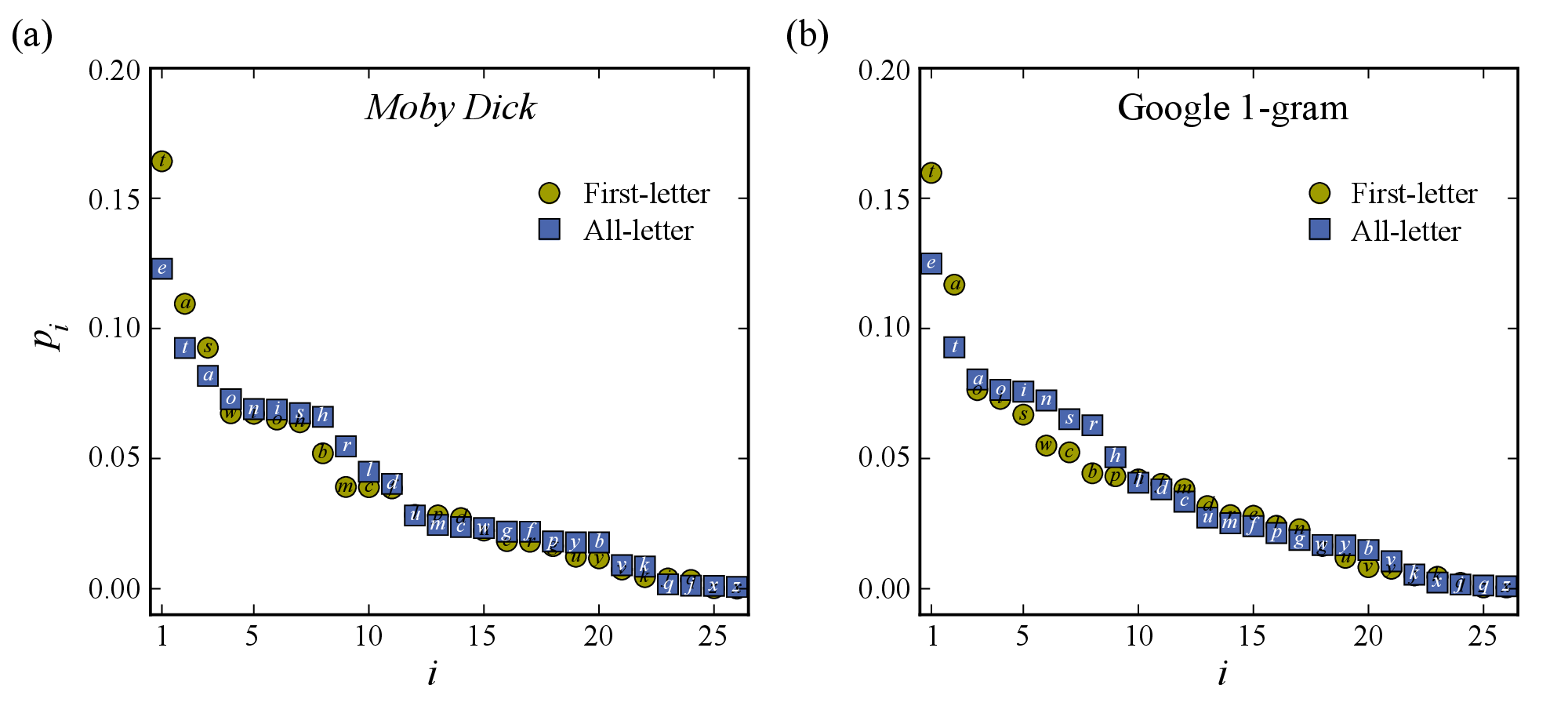}
\caption{
Comparison between the distribution of \textit{all-letters} (filled squares) with the distribution of \textit{first-letters} (filled circles). (a) for the novel {\it Moby Dick}. (b) for English 1-gram. Rank numbers 10-26 agree fairly well but the lowest rank numbers are different.
}
\label{fig9}
\end{figure*}

Our derivation of FLL uses the randomness of words in a novel, which carries over to a randomness of first letters. If you instead investigated the distribution of all letters, then the applicability of this particular argument is questionable. Figure~\ref{fig9} compares the distribution of \textit{first-letters} and \textit{all-letters} for {\it Moby Dick} and English 1-grams. One notices a good agreement for ranks 10-26 which deteriorates and becomes bad for the highest ranks 1-3. One also notices that the ranking in terms of letters is very different between \textit{all-letters} and \textit{first-letters}: for {\it Moby Dick} rank 1-3 are `{\it e}', `{\it t}', `{\it a}' and `{\it t}',`{\it a}',`{\it s}', respectively. Furthermore rank 1 `\textit{e}' for \textit{all-letters} has rank 16 for \textit{first-letters} and rank 4 `\textit{w}' for {\it first-letters} has rank 18 for \textit{all-letters}. Our conclusion is that even if there are similarities between the \textit{first-letters} distribution and \textit{all-letters} distribution, the striking agreement between FLL displayed in Fig.~\ref{fig3} only holds for the \textit{first-letters}.

\begin{figure}
\includegraphics[width=0.5\textwidth]{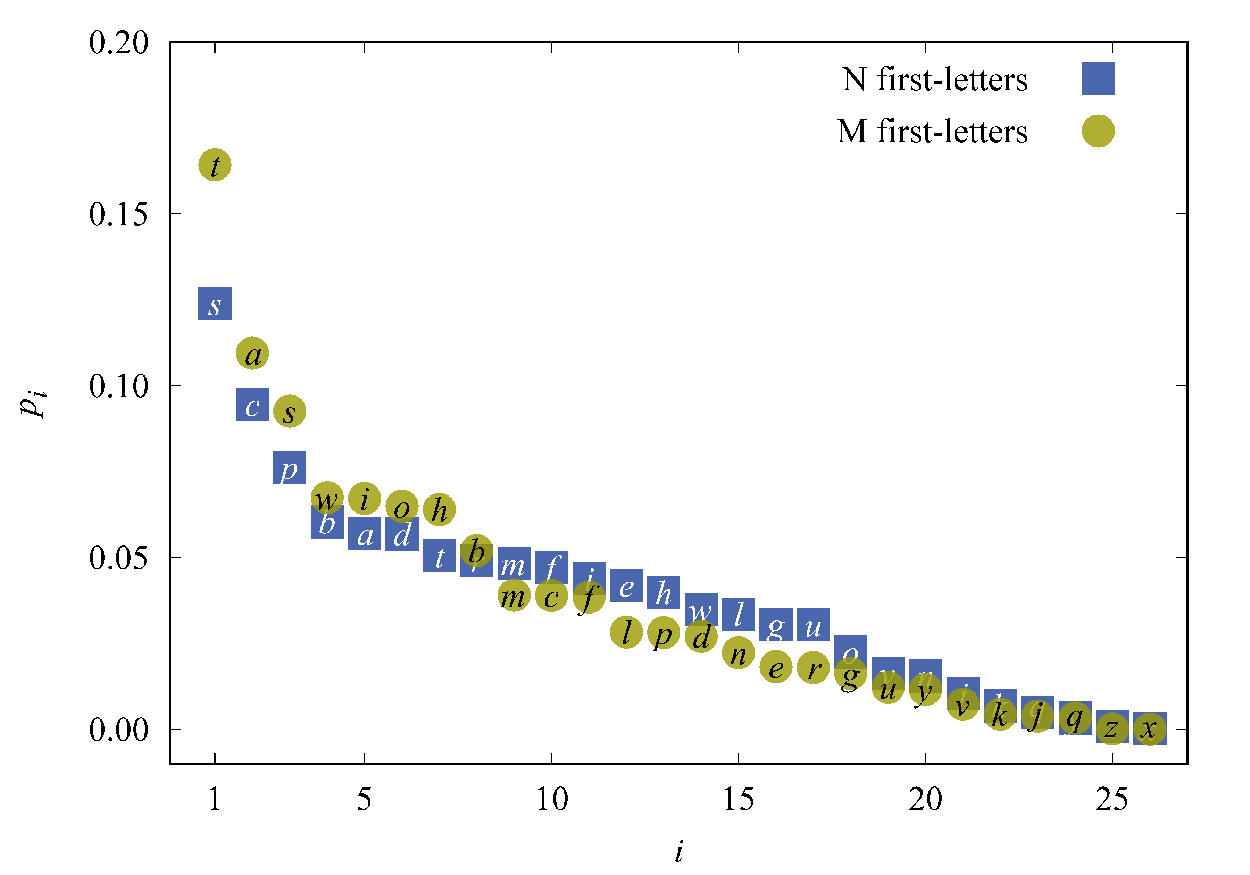}
\caption{
Comparison between the distribution of \textit{N first-letters} (filled squares) with the distribution of \textit{M first-letters} (filled circles) for the novel {\it Moby Dick}. Note that the ranking of the letters are quite different for the two cases.
}
\label{fig10}
\end{figure}

In Ref.~\cite{shulzinger2017} a different type of first letter distribution were discussed: the first letter distribution for a novel in the present investigation is based on the first letter of the total words $M$ of the novel. Reference~\cite{shulzinger2017} investigates the number of first letter words in dictionaries. This corresponds to the first letter of the $N$ distinct words in a novel. In Fig.~\ref{fig10} we compare our \textit{M first-letter} distribution with the corresponding \textit{N first-letter} in the case of {\it Moby Dick}.

The difference between \textit{M first-letter} and \textit{N first-letter} is readily understood: About 50\% of the words in an English novel, like {\it Moby Dick}, only occurs once in the text. The frequency of first letters for these 50\% corresponds to their occurrences in a dictionary. However, for the other 50\% the frequency of \textit{M first-letter} exceeds the corresponding frequency of \textit{N first-letter}: the more times a particular word occur in the text the larger becomes the difference. The most frequent word in an English text is `\textit{the}' which constitutes about 5\% of all the words. Thus for the first letter the difference should be expected to be large. As seen from Fig.~\ref{fig10}, this is indeed the case: 16.4\% words in total start on `\textit{t}', but only about 5\% specific words. Thus the difference between the distributions is due to a real difference and only partly to statistical fluctuations. This real difference is also reflected in the fact that the FLL predicts \textit{M first-letter} distributions much better than first letter frequencies obtained directly from a dictionary.

It is tempting to speculate that the existence of a universal frequency pattern in a language might have specific linguistic or cognitive implications. However, our belief is, just as for first digits and Benford's law case, that the phenomena is more general. To our mind it is in parallel with the findings in Ref.~\cite{yan16} that many situations which involve free human decisions, are nevertheless on the average moulded by a maximum entropy principle.

In summary: the present paper focuses on 
(i) the existence of a First-Letter Law, which is demonstrated from data, (ii) an explicit expression for this  universal law which gives the frequency-ladder of first letters, and (iii) The structural similarity with Benford's law, which is pointed out.

Even within limited scope one might argue that the present paper opens up for more questions than it answers. Such question might then be the subject of future research. 

\section*{Acknowledgments}
S.G.Y. and B.J.K. were supported by Basic Science Research Program through the National Research Foundation of Korea funded by the Ministry of Science, ICT and Future Planning with grant No. NRF-2017R1A2B2005957.

\begin{appendix}
\section{Derivation of First-Letter Law from exponential distribution}
\label{app:appA}
\setcounter{figure}{0}
\renewcommand{\thefigure}{A.\arabic{figure}}
\setcounter{equation}{0}
\renewcommand{\theequation}{A.\arabic{equation}}

The starting point is a collection of $M$ objects randomly assembled into $N$ groups of different sizes $k$. The probability for an object to be found within a group which contains objects which contains within the range $[k,\overline{k}]$ is given by $\int_k^{\overline{k}} P(k)dk$. In the present case $P(k)\propto \exp(-bk)$. The total number of groups is $N$ and consequently $N\int_k^{\overline{k}} P(k)dk$ is the expected average number of groups within the size interval $[k,\overline{k}]$. Each value $k$ is associated with a group which means that the largest group contains the $k$-values within the interval $[k_{c1}, \infty]$ where $k_{c1}$ is determined from the condition $N\int_{k_{c1}}^\infty P(k)dk=1$~\cite{baek11}. The average $k$-value for an object within this interval is given by 
\begin{equation}
k_1=\frac{\int_{k_{c1}}^\infty kP(k)dk}{\int_{k_{c1}}^\infty P(k)dk}.
\end{equation}
The average value $k_1$ is identified as the number of objects belonging to the largest group~\cite{baek11}. In the same way the value $k_2$ is identified as the size of the second largest group and so on. Thus the size of the group with size rank $i$ is given by 
\begin{equation}
k_i=\frac{\int_{k_{ci}}^{k_{c(i-1)}}kP(k)dk}{\int_ {k_{ci}}^{k_{c(i-1)}} P(k)dk}.
\end{equation}
For $P(k)\propto \exp(-bk)$ the cummulant $C(k)=\int_k^\infty P(k)dk$ is also an exponential and since $C(k=0)=1$ by the normalization condition for $P(k)$, it is given by $C(k)=\exp(-bk)$. The condition for $k_{ci}$ hence reduces to  $\exp(-bk_{ci})=i/N$ or $bk_{ci}=-\ln(i/N)$. The integral for $k_i$ gives
\begin{eqnarray}
k_i&=&\frac{\int_{k_{ci}}^{k_{c(i-1)}}k \exp(-bk)dk}{\int_{k_{ci}}^{k_{c(i-1)}}\exp(-bk)dk} \\
&=&\frac{1}{b} +\frac{k_i\exp(-bk_{ci})-k_{c(i-1)}\exp(-bk_{c(i-1)})}{\exp(-bk_{ci})-\exp(-bk_{c(i-1)})}.
\end{eqnarray}
and inserting the value $bk_{ci}$ reduces this to 
\begin{equation}
k_i=\frac{1}{b}\Bigl[1 +\ln N- i\ln i+(i-1)\ln(i-1)\Bigr].
\end{equation}
This gives the differences between the sizes of groups of different ranks as
\begin{equation}
 k_i-k_j\propto -i\ln i+(i-1)\ln(i-1)+j\ln j-(j-1)\ln(j-1),
\end{equation}
or
\begin{equation}
 k_i-k_j\propto -i\log_N i+(i-1)\log_N(i-1)+j\log_N j-(j-1)\log_N(j-1),
\end{equation}
since $\log_N x\propto \ln x$. Thus if you view $k_i-k_j$ as the height-difference between step $i$ and $j$ in a staircase, then the ratio of the height difference between these two steps and and two other steps $l$ and $m$ is given by   
\begin{eqnarray}
&\frac{k_i-k_j}{k_l-k_m}=\\
&\frac{-i\log_N i+(i-1)\log_N(i-1)+j\log_N j-(j-1)\log_N(j-1)}{-l\log_N l+(l-1)\log_N(l-1)+t\log_N m-(m-1)\log_N(m-1)}.
\end{eqnarray}
Thus the ratio of the step-height differences depends only on the total number of steps and the rank of the steps. The most general staircase compatible with these particular step-ratios is 
\begin{equation}
k_i \propto {\rm const.}-i\log_N i+(i-1)\log_N(i-1).
\label{AII}
\end{equation}
The group size ratio is given by $p_i=k_i/(\sum_{i=1}^Nk_i)$ with $k_i$  given by Eq.~(\ref{AII}). Furthermore
\begin{equation}
\sum_{i=1}^Nk_i \propto N\cdot {\rm const.} -\log_N N! +\sum_{i=2}\log_N\Bigl({\frac{i}{i-1}}\Bigr)^{(i-1)},
\end{equation}
which reduces to 
\begin{equation}
\sum_{i=1}^N k_i\propto N\cdot {\rm const.} -N=N\cdot {\rm const.}.
\end{equation}

In the case of alphabets $N=X$ where $X$ is the number of letters and the lowest rank usually has a vanishing or very small size. Requiring that $k_X=0$ results in
First-Letter Law given by Eq.~(\ref{alaw}). 

\section{}
\label{app:appB}
\setcounter{figure}{0}
\renewcommand{\thefigure}{B.\arabic{figure}}
\setcounter{equation}{0}
\renewcommand{\theequation}{B.\arabic{equation}}

\begin{figure*}
\includegraphics[width=1\textwidth]{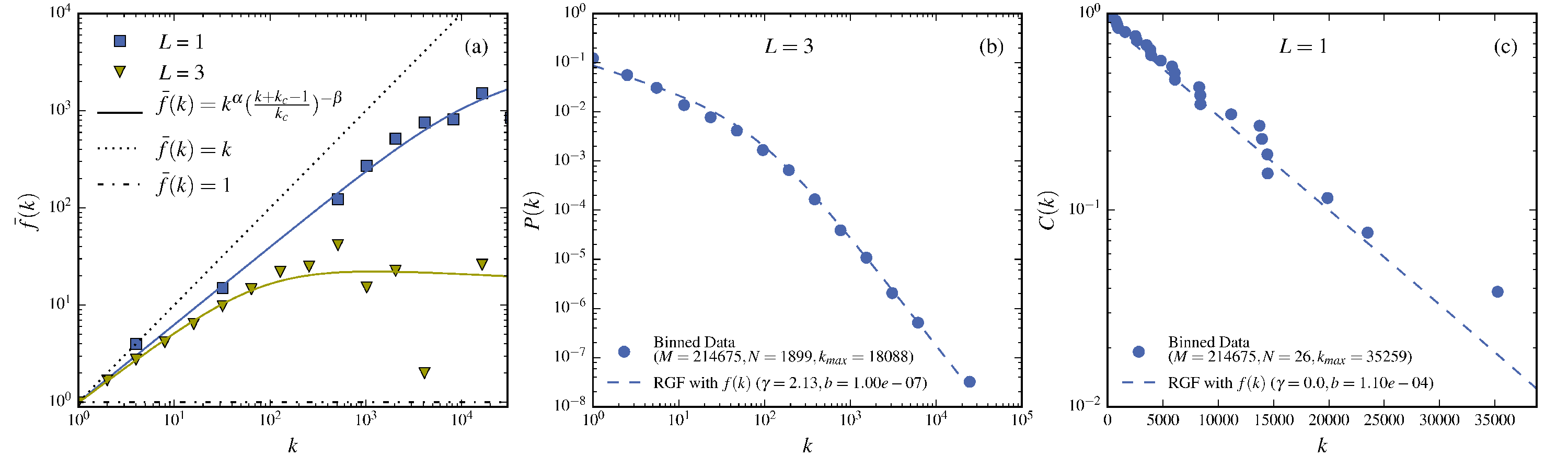}
\caption{
Relation between deletion of letters from words and loss of information. A
specific of meaning of a word in a text may be lost when the word is
represented by the first three or only the first letter because the three and
first letter representation of a word often corresponds to many words with
single meanings. The multiple meanings of these shortened words correspond to
an information loss. (a) shows this information loss for the novel {\it Moby
Dick}  plotted as the average meanings of shortened words which appears $k$
times in the text. Filled triangles corresponds to representations by the first
three letters and filled squares to only the first letters. Full straight line
corresponds to the extreme case when the representation of a word carries no
distinguishable meaning. The first letter representation is closing on the
straight line signaling a substantial information loss. (b) and (c) show that
the frequency distributions of the shortened words can be obtained from the
maximum entropy estimate given by the RGF theory: (b) compares this
RGF-prediction (broken curve) with the data (filled circles) for the first
three letters representation and (c) for only the first letter representation.
The agreements imply that the connection between information loss and the form
of the distribution is correct. (c) shows that the first-letter distribution is
close to an exponential.
}
\label{figB1}
\end{figure*}

Reference~\cite{yan17} describes how the word-frequency distribution changes shape as you consecutively delete letters from words. A short resume is given here: The word-frequency for {\it Moby Dick} is given in Fig.~\ref{fig8}(a). The broken curve gives the prediction from the maximum entropy argument given by the RGF(random group formation)-approximation~\cite{baek11,yan15,yan16,yan17}. The assumption made is that you randomly assemble $M$ labeled object into $N$ groups. $P(k)$ is then the average value of $N(k)/N$ where $N(k)$ is the number of groups which contain $k$ objects. The information needed to identify an object within a group of $k$ objects is $\log_2 k$, so that the average information needed to identify a particular objects is $\sum_kP(k)\log_2[kN(k)]$ which up to a constant is $\sum_kP(k)\log_2[kP(k)]$.
The maximum entropy in the RGF-approach corresponds to the minimum of the average information $\sum_kP(k)\log_2[kN(k)]$ subject to appropriate constraints. If the only constraints are that $M$ and $N$ are known, then the solution is $P(k)\propto \exp(-bk)/k$. In the limit of $M\gg N$ this reduces to $\propto 1/k$.

 In the RGF-approach you are looking for the minimum of the information which for given $M$ and $N$ in addition has a fixed value of entropy. This changes the solution to $P(k)\propto \exp(-bk)/k^{\gamma}$. This means that for a given values of $M$, $N$, and $k_{\rm max}$ (frequency of the most frequent word), you obtain a unique prediction. This prediction in case of {\it Moby Dick} is shown in Fig.~\ref{fig8}(a). 

One interpretation of this is that the entropy constraint effectively changes the information needed to identify an object in a group of size $k$ from $\log_2 k$ to $\gamma \log_2 k$. This connection between information and word-frequency was in Ref.~\cite{yan17} explored by deleting letters from words. When you delete letters from a word it may acquire multiple meanings. Figure~\ref{figB1}(a) shows the average number of  meanings, $f(k)$, of a word which occurs $k$-times in the text will acquire if you keep only the first letter of a word ($L=1$) or the first three letters ($L=3$). This means that your capacity of distinguishing between words has deteriorated from $\log_2 k$ to $\log_2[k/f(k)]$~\cite{yan15,yan16,yan17}. Your average information becomes $\sum_kP(k)\log_2[kP(k)/f(k)]$ and the RGF-approach for this information gives the predictions given in Fig.~\ref{figB1}(b) and (c). As you keep less letters you gradually lose the capability of distinguishing between words. In case of keeping only the first letter, RGF-prediction gives $P(k) \propto \exp(-bk)$ to very good approximation. This means that the capability of distinguishing between words has been lost almost altogether. It corresponds to the situation when $f(k)=k$ and hence the reduced information $\log_2(k/k)=0$. In other words, the limitation case is when the words are completely indistinguishable. Thus the closer a text is to this limiting case, the better described by the First-Letter Law.
\end{appendix}


\begin{thebibliography}{99}
\bibitem{miller2015} S.J. Miller (Ed.), Benford's Law: Theory and Applications, first ed., Princeton University Press,
Princeton and Oxford, 2015.

\bibitem{census}[dataset] United States Census Bureau, 2013 US population census data, 2013. http://www.census.gov/population/.

\bibitem{benford1938}
F. Benford, The Law of Anomalous Numbers, Proc. Am. Philos. Soc. 78 (1938) 551-572.

\bibitem{newcomb1881}
S. Newcomb, Note on the Frequency of Use of the Different Digits in Natural Numbers, Amer. J. Math 4 (1881) 39-40.

\bibitem{pietronero2001}
L. Pietronero, E. Tosatti, V. Tosatti, A. Vespignani, Explaining the uneven distribution of numbers in nature: the laws of Benford and Zipf, Physica A 293 (2001) 297-304.

\bibitem{whyman2016}
G. Whyman, E. Shulzinger, Ed. Bormashenko, Intuitive consideration clarifying the origin and applicability of the Benford law, Result Phys. 6 (2016) 3-6.

\bibitem{fewster2009} R.M. Fewster, A Simple Explanation of Benford's Law, Am. Stat. 63 (2009) 26-32.

\bibitem{gutenberg}[dataset] Gutenberg Project, Novel data, http://www.gutenberg.project.

\bibitem{google}[dataset] Google Ngram, 1-grams data, version 20120701, 2012. http://storage.googleapis.com/books/ngrams/books.

\bibitem{baek11}
S.K. Baek, S. Bernhardsson, P. Minnhagen, Zipf's law unzipped, New J. Phys. 13 (2011) 043004.

\bibitem{yan15} 
X. Yan, P. Minnhagen, Maximum entropy, word-frequency, Chinese characters, and multiple meanings, PLoS ONE 10 (2015) e0125592.

\bibitem{yan15b}
X. Yan, P. Minnhagen, Randomness versus specifics for word-frequency distributions, Physica A, 444 (2016) 828-837.

\bibitem{yan16}
X. Yan, P. Minnhagen, H.J. Jensen, The likely determines the unlikely, Physica A 456 (2016) 112-119.

\bibitem{yan17} 
X. Yan, P. Minnhagen, The Dependence of Frequency Distributions on Multiple Meanings of Words, Codes and Signs, submitted to Physica A.

\bibitem{lee12}
S.H. Lee, S. Bernhardsson, P. Holme, B.J. Kim, P. Minnhagen, Neutral theory of chemical reaction networks, New J. Phys. 14 (2012) 033032.

\bibitem{shulzinger2017}
E. Shulzinger, Ed. Bormashenko, On the universal quantitative pattern of the distribution of initial characters in general dictionaries: the exponential distribution is valid for various languages, J. Quant. Linguist. (2017) 1-16.

\end{thebibliography}
\end{document}